\pdfoutput=1

\documentclass[11pt]{article}

\usepackage{authblk}

\usepackage[]{EMNLP2023}

\usepackage{times}
\usepackage{latexsym}

\usepackage{graphicx}

\usepackage[T1]{fontenc}

\usepackage[utf8]{inputenc}

\usepackage{microtype}

\usepackage{inconsolata}

\usepackage{amsmath}

\usepackage{caption}
\usepackage{subcaption}

\usepackage{multirow}
\usepackage{rotating}

\usepackage{tcolorbox}
\tcbuselibrary{listings,breakable,skins}

\title{Don’t Add, don’t Miss: Effective Content Preserving Generation from Pre-Selected Text Spans
}

\author[1]{\bf Aviv Slobodkin}
\author[1,2]{\bf Avi Caciularu}
\author[1]{\bf Eran Hirsch}
\author[1]{\bf Ido Dagan}
{
\makeatletter
\renewcommand\AB@affilsepx{~~~ \protect\Affilfont} \makeatother
\affil[1]{Bar-Ilan University}
\affil[2]{Google Research}
}
\affil[  ]{} 
\affil[  ]{\tt \{lovodkin93, hirsch.eran\}@gmail.com}
\affil[  ]{\tt avica@google.com} 
\affil[  ]{\tt dagan@cs.biu.ac.il}

\begin{document}
\maketitle
\begin{abstract}
    The recently introduced Controlled Text Reduction (CTR) task isolates the text generation step within typical summarization-style tasks. It does so by challenging models to generate coherent text conforming to pre-selected content within the input text (``highlights''). 
    This framing enables increased modularity in summarization-like tasks, allowing to couple a single CTR model with various content-selection setups and modules. 
    However, there are currently no reliable CTR models, while the performance of the existing baseline for the task is mediocre, falling short of practical utility.
    Here, we address this gap by introducing a high-quality, open-source CTR model that tackles two prior key limitations: inadequate enforcement of the content-preservation constraint, and suboptimal silver training data. 
    Addressing these, we amplify the content-preservation constraint in both training, via RL, and inference, via a controlled decoding strategy. 
    Further, we substantially improve the silver training data quality via GPT-4 distillation. 
    Overall, pairing the distilled dataset with the highlight-adherence strategies yields marked gains over the current baseline, of up to 30 ROUGE-L points, providing a reliable CTR model for downstream use.\footnote{Our code is publicly available at \url{https://github.com/lovodkin93/CDR_CTR}}
\end{abstract}

 \begin{figure*}[ht!]
\centering
    \includegraphics[width=\textwidth]{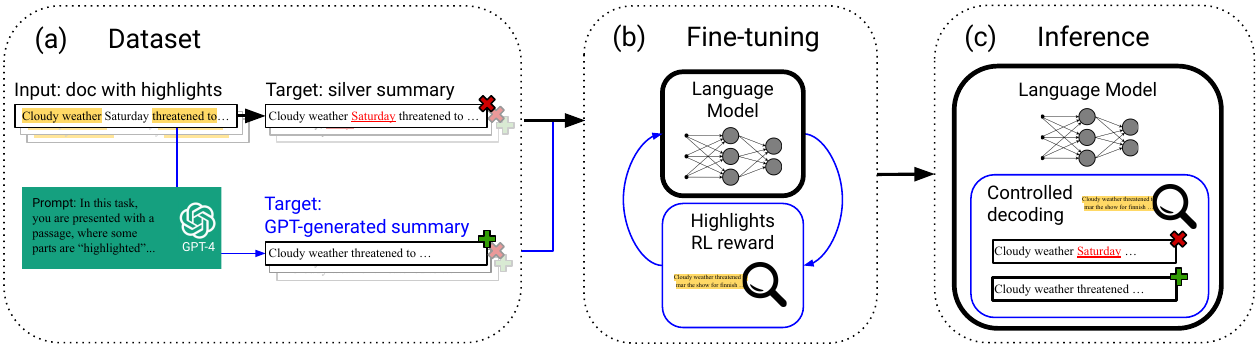}
    \caption{Overview of our contributions, encompassing three modeling phases.  Components introduced in our approach are denoted in blue. (a) We generate new target summaries using GPT-4, conditioned on the silver highlights in the original dataset. (b) During training, we fine-tune our model taking an RL approach, based on Quark \citep{lu2022quark}. (c) During inference, we employ a highlights-centric controlled decoding algorithm.}
    \label{fig:full-process}
\end{figure*}


\section{Introduction}\label{sec:Introduction}
\begin{figure*}[ht!]
\centering
    \includegraphics[width=\textwidth]{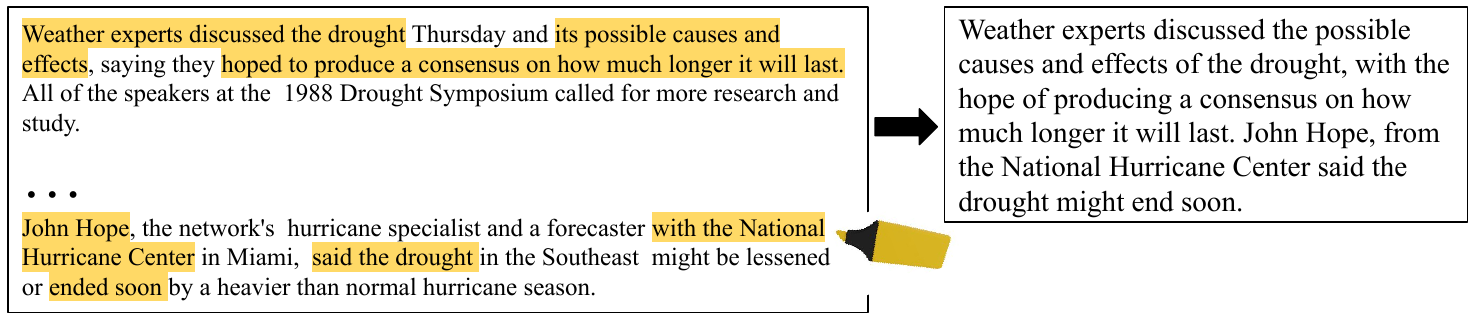}
    \caption{Demonstration of the Controlled Text Reduction task. The input consists of a source document and highlights (left), and the desirable output covers exclusively the highlighted content while preserving coherence (right). Borrowed and adapted from \citet{slobodkin-etal-2022-controlled}.}
    \label{fig:CTR_demonstration}
\end{figure*}

The abstractive text summarization task, aiming to generate accurate and coherent summaries from one or multiple documents, involves two principal sub-tasks: (a) identification of salient information in the input text(s) and (b) its consolidation into a coherent text.
Recently, \citet{slobodkin-etal-2022-controlled} proposed an explicit decomposition of these two subtasks, particularly concentrating on the latter as an isolated task termed \textit{Controlled Text Reduction} (CTR), as illustrated in 
 \autoref{fig:CTR_demonstration}.
This task takes as input a text with pre-selected marked spans (``highlights'') and expects a reduced and coherent version of the text, covering precisely the content of these input spans. 
In addition to a baseline model, the authors also provided crowdsourced dev and test sets, along with automatically-generated silver training data, where each instance comprises of a document with highlighted content and the corresponding reduced text summary.
The proposed adoption of CTR offers greater control over text generation, enabling modular summarization systems, where a single CTR model can be combined with various content selection strategies and user preferences. 
For example, the same  CTR model could be used for generation from content selected for either generic or query-focused summarization \citep[QFS;][]{dang-2006-duc}, or for long-form question-answering \citep[LFQA;][]{fan-etal-2019-eli5}. It could even enable a human-in-the-loop scenario, where customized summaries can be created, based on users' preferences, as was recently demonstrated in \citet{slobodkin-etal-2023-summhelper}.
Further, by excluding the \textit{subjective} content selection requirement of the full summarization task, CTR offers a semantically well-defined and objective generation task, focused on coherent content consolidation.

Being recently introduced, there currently exists no high-quality CTR model, with the present baseline succeeding to cover only half of the highlighted details while pulling much non-highlighted content from the surrounding context. 
In this paper, we aim to design an efficient CTR model of sufficient quality for reliable integration in modular architectures.
Our proposed method, outlined in \autoref{fig:full-process}, addresses two different shortcomings of the existing resources.
First, we examine methods to intensify the highlights signal during training and inference, in order to yield better content preservation. Second, we address the noise evident in the available silver training data, improving its quality using GPT-4.

We begin our investigation by exploring strategies for enforcing the highlights signal. First, we explore the use of Reinforcement Learning (RL) during \textit{training} to both accentuate the highlights signal and mitigate the impact of the data noise, adapting the recently-introduced Quark algorithm \citep{lu2022quark}, which aims to \textit{un}learn unwanted properties.
Since RL methods bias the model to conform to a reward function, in addition to the training data, we hypothesize it has the potential to reduce the impact of noisy data biases on the model's behavior.
We then design a highlight-aware decoding mechanism, following \citet{wan-etal-2023-faithfulness}, which 
biases the model to better preserve the highlights content at \textit{inference} time.
Finally, we address the inherent noise within the available silver training data, by employing GPT-4 \citep{openai2023gpt4} to generate cleaner training data for our model, essentially performing (symbolic) distillation from GPT-4.

Empirically, we demonstrate that each of the aforementioned strategies separately yields state-of-the-art results, surpassing the baseline model in terms of highlights content preservation.
Further, we show that GPT-4 is indeed effective in generating better silver training data, leading to further improvements.

Hence, our contribution in this paper is twofold:
\begin{enumerate}
    \item Proposing and investigating multiple strategies to amplify the highlights signal in the CTR setting, addressing training, inference, and data generation.
    \item Developing a high-quality CTR model, significantly outperforming the available baseline.
\end{enumerate}


\section{Background}\label{sec:Background}

This section provides the needed background regarding our task and the methods we employ.

\paragraph{Controlled Text Reduction}\label{subsec:Background-CTR}
Controlled Text Reduction \citep[CTR;][]{slobodkin-etal-2022-controlled} is a recently introduced task that aims to generate a reduced version of a text that \textit{exactly} covers pre-determined selected content, referred to as ``highlights'' (see \autoref{fig:CTR_demonstration}). It effectively generalizes the sentence de-contextualization task \citep{choi-etal-2021-decontextualization}, which
addresses only the case of rephrasing a single full sentence given in context to be comprehensible standalone.
In contrast to the full summarization task, which involves a substantial degree of subjectivity in content selection, CTR requires exact preservation of modularly \textit{pre-selected} content, making the task more semantically objective. At the same time, this stringent requirement for both faithfulness to and coverage of the highlights, makes the task more semantically challenging than standard summarization, which is obliged only to faithfulness, but not to full coverage.

Accompanying the task, the CTR authors introduced a manually annotated development and test sets, as well as an automatically-generated silver train set, derived from the DUC summarization dataset\footnote{\url{https://duc.nist.gov/}} using a summary-source alignment model \citep[SuperPAL;][]{ernst-etal-2021-summary}. 
However, an approximate $30\%$ mismatch surfaced between the `silver highlights', namely source spans identified by SuperPAL as related to the summary, and their respective summary content. Such discrepancy may cause a model trained on this data to develop biases toward overlooking certain types of highlights during training, as well as including salient, yet non-highlighted content.
Here, we address this issue by improving the CTR training dataset.



\paragraph{Controlling via Reinforcement Learning}
\label{paragraph:bg-RL}
Reinforcement Learning (RL) has been increasingly utilized to control various facets of text generation \citep{pasunuru-bansal-2018-multi, yuan-etal-2019-interactive, https://doi.org/10.48550/arxiv.2112.09332}, with most works relying either on REINFORCE \citep{williams1992simple}, an algorithm notable for its direct yet high-variance approach to policy optimization, or the Proximal Policy Optimization \citep[PPO;][]{schulman2017proximal}, an algorithm renowned for efficiently balancing policy stability and learning. There has also been a growing interest in using RL methods for reducing undesired behaviors, including toxicity \citep{Faal_2022} and redundancy \citep{mao2020multi}.

Following this line of work, \citet{lu2022quark} recently introduced Quark - an algorithm inspired by the Proximal Policy Optimization algorithm, designed to \textit{un}learn unwanted properties, which we leverage here. 
While REINFORCE and PPO are used to learn a desired policy, Quark aims to remove or reduce specific behaviors from the learned policy, enabling it to effectively address the undesired behaviors in the text generation process.
The algorithm iteratively alternates between three steps: (1) \textbf{Exploration}, where the current model state generates new input-output samples from the training inputs, and then incorporates them into the stored data pool; (2) \textbf{Quantization}, where a predefined reward function is used to rank and classify the accumulated data pool into $K$ quantiles, with a reward token assigned to each; and (3) \textbf{Learning}, where the algorithm maximizes the likelihood of the accumulated data pool, with the instances conditioned on their quantile labels. A KL-divergence penalty is also applied during training to ensure proximity to the original language model distribution \citep{https://doi.org/10.48550/arxiv.1611.02796, ziegler2019finetuning}.

The objective of Quark is to teach the model to generate texts of varying quality with respect to the reward function. Then, at inference, the model is asked to generate high-reward outputs. The algorithm exhibits state-of-the-art results in several attribute-removal objectives, such as toxicity and unwanted sentiment, surpassing many other RL-based text generation approaches, including PPO.

In this work, we propose modifying Quark to teach models to unlearn biases caused by the somewhat noisy training data, thereby enhancing their performance in adhering to the highlighted content. 


\paragraph{Controlled Decoding}\label{paragraph:bg-controlled_decoding}
Controlled decoding algorithms aim to guide models to better address concrete output requirements that can be measured and enforced during decoding.
To this end, various works designed constraint-sensitive decoding methods that modify the search space in accordance with the constraints \citep{anderson-etal-2017-guided, hokamp-liu-2017-lexically, post-vilar-2018-fast, lu-etal-2021-neurologic, slobodkin-etal-2023-curious}. 

Recently, a faithfulness-aware decoding mechanism was proposed by \citet{wan-etal-2023-faithfulness}, which involves a lookahead operation during decoding, inspired by \citet{lu-etal-2022-neurologic}. At every decoding step, the algorithm projects into the future to create a complete summary commencing with the current tokens of any partially formed summary. It then selects tokens that provide paths exhibiting enhanced faithfulness within the search space. 
Formally, each token's score is calculated by:
\begin{equation}\label{eq:CTR_decoding}
f(y_t) = logP(y_{\leq t}|x) + \lambda\cdot\max_{\mathcal{L}_l(y_{\leq t})} g(y_{\leq t+l}, x)
\end{equation}
where $x$ represents the current input tokens, $t$ is the generation step, $y$ is the generated output, $logP(y_{\leq t}|x)$ is the underlying generation score, $g(\cdot)$ is a faithfulness evaluation function, $\lambda$ stands as a hyperparameter to regulate the emphasis placed on future predictions, and $l$ stands for the number of tokens to look into the future. $\mathcal{L}_l(y_{\leq t})$ is a collection of length $l$ continuations of $y_{\leq t}$, and its size ranges from a single continuation in a greedy decoding approach, to $k$ potential continuations in beam-search decoding.
In this work, we adapt this algorithm, shifting its focus towards precise matching with the highlighted content rather than being (only) faithful to the entire input.

\paragraph{Data Generation with Language Models}
Recently, many works proposed using LM-generated data to directly train smaller manageable models. These efforts address various challenges, including controllability \citep{sclar2022referee}, model reasoning \citep{zelikman2022star, hsieh2023distilling}, and language understanding \citep{ye-etal-2022-zerogen, han2022unsupervised}. These works align with the Symbolic Knowledge Distillation scheme \citep{west-etal-2022-symbolic}, where knowledge from the teacher model is transferred via a textual dataset used for student training.
Here, we employ GPT-4 to generate improved silver training data for our models.

\section{Method}\label{sec:Methods}
This section presents the three core components of our contribution: highlights-driven reinforcement learning (RL) (\S\ref{subsec:methods-Highlights-Oriented RL Training}), highlight-attentive controlled decoding (\S\ref{subsec:methods-Highlights-Sensitive Decoding}), and the steps taken for generating improved fine-tuning data with GPT-4 (\S\ref{subsec:Distillation from GPT-4}), corresponding to components (b), (c), and (a) in \autoref{fig:full-process}, respectively.

\subsection{Highlights-Oriented RL Training}
\label{subsec:methods-Highlights-Oriented RL Training}
To adapt Quark (see \S\ref{paragraph:bg-RL}) for the CTR task, we introduce a highlights-focused reward, based on the ROUGE metric \citep{lin-2004-rouge}.\footnote{We also experimented with PPO, but adopted Quark which outperformed it on the development set.} 
Previous RL-based text generation studies with ROUGE-based rewards calculated ROUGE scores relative to the gold reference summaries. In contrast, we propose calculating the ROUGE rewards for the generated output compared to the concatenated \textit{input highlights}, to encourage their content preservation.
Furthermore, to motivate the model to balance between the task's two requirements, namely, covering the entire highlights and avoiding the inclusion of excessive non-highlighted content, we suggest optimizing each of these objectives separately.
Inspired by the dual-reward procedure \citep{pasunuru-bansal-2018-multi}, we propose alternating between two highlights-focused rewards:
One that encourages coverage of highlights, for which we use ROUGE \textit{recall}, and another that prioritizes adherence (faithfulness) to the highlights, for which we employ ROUGE \textit{precision}. Indeed, we find that this alternating reward strategy works best over the development set (see Appendix~\ref{sec:reward_tuning}).

\subsection{Highlights-Sensitive Decoding}
\label{subsec:methods-Highlights-Sensitive Decoding}
To bias the model to better preserve the highlights content at inference time, we follow the faithfulness-aware decoding strategy from \citet{wan-etal-2023-faithfulness} (see \S\ref{paragraph:bg-controlled_decoding}). Here, we adapt this method to prioritize partially formed summaries that are likely to eventually (once completed) match better the pre-selected highlights. To that end, we substitute the score $g(y_{\leq t+l}, x)$ in \autoref{eq:CTR_decoding} with a new score, $g(y_{\leq t+l}, x_{h})$, where $x_{h}$ represents the concatenation of the highlights. In essence, our strategy shifts from an input-focused to a highlights-focused scoring approach, while requiring both faithfulness and complete coverage.
Within this framework, we evaluate two potential metrics for the computation of $g(y_{\leq t+l}, x_{h})$: ROUGE-L F1 and METEOR scores.
We found that ROUGE-L F1 consistently matched or exceeded METEOR's performance, and hence adopted it for our method. Please refer to Appendix~\ref{sec:Highlights-Focus Controlled Decoding Hyperparameter Tuning}  for further details. 

\subsection{Improving Silver Training Data with GPT-4}\label{subsec:Distillation from GPT-4}
We wish to improve the quality of the CTR dataset, by leveraging the capabilities of GPT-4 \citep{openai2023gpt4}.  
The existing CTR training dataset consists of summaries manually composed by experienced summarizers, while the highlights were automatically identified using a summary-source alignment model (see \S\ref{subsec:Background-CTR}). As discussed, the performance of this alignment model leaves much room for improvement.

To capitalize on GPT-4's abilities in text generation, we employ it to generate more fitting summaries, based on the silver highlights. To that end, we supply GPT-4 with a modular prompt, inspired by the Chain-of-Thought approach \citep{wei2023chainofthought}. 
This custom prompt incorporates two exemplars that deconstruct the controlled reduction into three steps: (1) the listing of highlights, (2) the consolidation of highlights on a sentence-by-sentence basis, and (3) the production of the ultimate reduction.
The detailed structure of this prompt can be found in Appendix~\ref{sec:GPT-4_Prompt}.

We will henceforth refer to models trained on the GPT-4-generated data as distilled, and to those trained on the original CTR data as non-distilled, following the notion of Symbolic Knowledge Distillation \citep{west-etal-2022-symbolic}.
As shown in Section~\S\ref{subsec:data_quality}, we observed an improved quality of the dataset, yielding better alignments between highlights and summaries.

\begin{table*}[t!]
\centering
\resizebox{0.75\textwidth}{!}{%
\begin{tabular}{|c|l|ccccc|c|}
\hline
& model & R-1 & R-2 & R-L & M & BertScore & Coherency \\ \hline
\multirow{5}{*}{\rotatebox[origin=c]{90}{\normalfont \hspace{-3mm} Non-Distilled}} & LED\textsubscript{H} (baseline) & 70.2 & 53.8 & 54.4 & 64.3 & 71.9 & 4.26 \\ \cline{2-8} 
& Flan-T5\textsubscript{H} & 74.1 & 59.1 & 63.4 & 67.4 & 73.2 & 4.56 \\
& \hspace{3mm}  {\small+ RL} & 75.6 & 63.1 & 68.9 & 73.6 & 76.8 & 4.34 \\
& \hspace{3mm}  {\small+ Con. Decoding} & \textbf{80.7} & \textbf{68.6} & \textbf{76.1} & 74.2 & \textbf{80.0} & 4.40 \\
& \begin{tabular}{@{}l} \hspace{3mm} \vspace{-1.5mm}  {\small+ RL}\\  \hspace{6mm}  {\small+ Con. Decoding} \end{tabular} & 78.1 & 66.4 & 73.4 & \textbf{75.1} & 76.8 & 4.42 \\ \hline
\multirow{4}{*}{\rotatebox[origin=c]{90}{\normalfont \hspace{-4mm} Distilled}} & Flan-T5\textsubscript{H} & 84.3 & 72.4 & 79.9 & 80.9 & 75.8 & 4.32 \\
& \hspace{3mm}  {\small+ RL} & 85.0 & 74.3 & 82.1 & \ \ \textbf{84.3$^*$} & 81.9 & 4.40 \\
& \hspace{3mm}  {\small+ Con. Decoding} & \ \ \textbf{88.3$^*$} & \ \ \textbf{79.1$^*$} & \ \ \textbf{86.4$^*$} & 84.0 & \ \ \textbf{84.2$^*$} & 4.54 \\
& \begin{tabular}{@{}l} \hspace{3mm} \vspace{-1.5mm} {\small+ RL}\\  \hspace{6mm}  {\small+ Con. Decoding} \end{tabular} & 86.7 & 76.4 & 84.6 & 84.1 & 81.9 & 4.44 \\ \hline
& GPT-4 & 82.3 & 67.1 & 75.1 & 79.2 & 80.8 & 4.58 \\ \hline
\end{tabular}%
}
\caption{ROUGE, METEOR (M), and BertScore results on the CTR testset, compared to the concatenated highlights, as well as coherency results. In addition to the baseline LED\textsubscript{H} and Flan-T5\textsubscript{H}, we also evaluate the combination of Flan-T5\textsubscript{H} with the highlights-sensitive decoding strategy ("+ Con. Decoding"), with the highlights-focused RL strategy ("+ RL"), and with their combination. We also evaluate all those variants when trained with the GPT-4-generated trainset ("Distilled") as well as GPT-4 itself, in a few-shot setting with two exemplars. For each metric, the best non-distilled and distilled Flan-T5 models are in bold, with the best overall model having an asterisk.}
\label{tab:rouge-results}
\end{table*}

\section{Experimental Setup}\label{sec:experimental_setup}
\paragraph{Base Model}

Throughout our experiments, our primary base model is the instruction-finetuned Flan-T5 large model \citep[Flan-T5\textsubscript{large};][]{https://doi.org/10.48550/arxiv.2210.11416}, further fine-tuned on the highlights-focused CTR dataset. 
The selection of this particular model is motivated by emergent research that indicates instruction-finetuned models manifest superior performance in tasks necessitating constrained generation \citep{sanh2022multitask, wei2022finetuned, zhou2023controlled}. We will refer to this model as Flan-T5\textsubscript{H}, where H stands for ``highlight-finetuned''.


In addition to this variant of Flan-T5\textsubscript{H}, we also show results of a large variant of the original pretrained CTR baseline model, which is the Longformer Encoder-Decoder large model \citep[LED\textsubscript{large};][]{https://doi.org/10.48550/arxiv.2004.05150}, finetuned on the CTR dataset. We will refer to this model as LED\textsubscript{H}.
Lastly, we also show the results of the few-shot GPT-4 \citep{openai2023gpt4}, when guided with the same prompt used in our distillation process (see \S\ref{subsec:Distillation from GPT-4}).

\paragraph{Highlights-Oriented RL Training}
Our highlights-sensitive RL training involves finetuning the anchor Flan-T5\textsubscript{H} model using Quark, combined with our highlights-driven reward policy (see \S\ref{subsec:methods-Highlights-Oriented RL Training}). Specifically, the first Exploration step (see \S\ref{sec:Background}) utilizes the pretrained and fine-tuned Flan-T5\textsubscript{H} to generate the initial pool of samples, which triggers the Quark unlearning process, performed on Flan-T5\textsubscript{H}. Following \citet{lu2022quark}, we set the number of quantiles to eight (see \S\ref{sec:Background}). 
We find, on the development set, that ROUGE-L is the most effective metric for our dual-reward function, yielding the greatest improvements relative to the ROUGE-1 and ROUGE-2 metrics (henceforth, ROUGE rewards refer to ROUGE-L).

\paragraph{Highlights-Sensitive Decoding}
To score the degree of alignment between each future completion of a current generated prefix and the highlights, we utilize the F1 measure of ROUGE-L, relative to the concatenated highlights (see \S\ref{paragraph:bg-controlled_decoding} and  \S\ref{subsec:methods-Highlights-Sensitive Decoding}). We use the hyperparameters used in \citet{wan-etal-2023-faithfulness}. Additionally, following its tuning, we chose a beam size of 8. For a more comprehensive discussion of the hyperparameter tuning, see Appendix~\ref{sec:Highlights-Focus Controlled Decoding Hyperparameter Tuning}.

 \begin{figure}[t!]
\centering
    \includegraphics[width=\columnwidth]{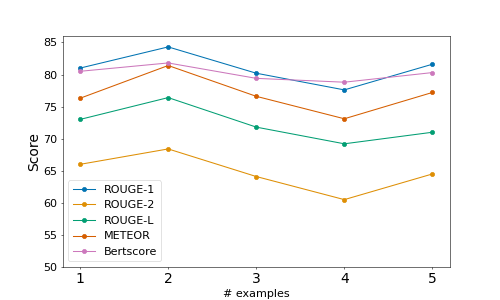}
    \caption{ROUGE, METEOR, and BertScore results on 50 instances from the CTR development set of GPT-4 models, for varying numbers of in-context examples in the prompt.}
    \label{fig:GPT4-number-examples-tuning}
\end{figure}

\paragraph{Generating New Training Data with GPT-4}
As part of our methodology to improve the generation of the silver training data, we evaluated the effect of varying the number of examples used in the few-shot prompt for GPT-4. This procedure involved the random sampling of fifty instances from the CTR development set, and their subsequent incorporation within a few-shot prompt. The quantity of in-context examples was systematically varied between one and five, the results of which are documented in \autoref{fig:GPT4-number-examples-tuning}. These results show that the use of two examples delivered the most favorable results across all evaluative metrics, thereby leading us to select this configuration in our experiments. 
We also experimented with ChatGPT and found it to be inferior to GPT-4, leading to the adoption of GPT-4 in this setting. For a more in-depth discussion of these findings and additional hyperparameter tuning, please refer to Appendix~\ref{sec:prompt_tuning}.

\paragraph{Evaluation}
For evaluation, we adopt the evaluation approach outlined by \citet{slobodkin-etal-2022-controlled}, calculating several content-matching metrics between the generated texts and the concatenated highlights. 
The rationale behind evaluating against the highlights, rather than the gold summaries, is rooted in CTR's principal requirement: an exact matching to the \textit{highlighted content},  not necessarily to the reference summary. While the highlights and reference summary are supposed to express the same content, some discrepancies may occur even in the gold test data.  Moreover, since automatic metrics are not perfect, models that are less abstract than the human-generated gold summaries, or that exhibit different paraphrastic abstractions, would unjustly be penalized when evaluated against the gold summaries. In Appendix~\ref{sec:qualitative_analysis}, we conduct a qualitative analysis that further shows the advantage of directly comparing outputs with the highlights, rather than with the gold summaries.

In addition to ROUGE \citep{lin-2004-rouge}, used in \citet{slobodkin-etal-2022-controlled}, we also measure METEOR scores \citep{denkowski-lavie-2014-METEOR}, informed by recent research underscoring the correlation of this metric with human judgment concerning relevance and consistency \citep{10.1162/tacl_a_00373}. We also report results on the  BertScore metric, which is less lexical and more semantic in nature, though not necessarily more reliable. 
Experimenting also with NLI-based metrics, we observed 
inadequate performance when applied over sub-sentence spans (our highlights), and hence leave it for future research to develop NLI-based metrics that are robust to such settings.

Finally, we also follow \citet{slobodkin-etal-2022-controlled}'s coherency analysis, by hiring crowd-workers to assess the coherency of 50 random samples for each model, with a 5-point Likert scale (for details see Appendix~\ref{sec:Fluency_Human_Annotation_Protocol}). Notably, our coherency analysis tested both coherency and fluency, following the settings in the original CTR paper. 
This approach follows the standard common practice, where fluency and coherence are best evaluated manually.
\section{Results and Analysis}\label{sec:Results}

\subsection{Non-distilled Models}

\autoref{tab:rouge-results} presents performance results, for models trained over the original CTR silver training data~\cite{slobodkin-etal-2022-controlled}). Primarily, Flan-T5\textsubscript{H} exhibits superior performance compared to LED\textsubscript{H}.
This trend is consistently apparent in all subsequent variants, hence, we confine our reporting of subsequent models to Flan-T5\textsubscript{H}'s variants.\footnote{See Appendix~\ref{sec:Results of All Variants with LED as the Backbone} for the results of all the variants with LED\textsubscript{H} as the backbone model.}

We observe that further finetuning Flan-T5\textsubscript{H} via our highlights-oriented RL protocol (``+ RL'' in \autoref{tab:rouge-results}) yields substantial improvements. Additionally, augmenting Flan-T5\textsubscript{H} with the highlights-aware decoding strategy (``+ Con. Decoding'' in \autoref{tab:rouge-results}) leads to an even larger performance improvement across ROUGE and BertScore metrics and a modest improvement in the METEOR metric. We find this sensible, given the aggressive nature of the controlled decoding approach in amplifying the highlight-signal when actually generating the output at inference time.
Interestingly, the incorporation of both strategies simultaneously results in a slight drop in performance based on the ROUGE and BertScore metrics compared to the controlled decoding variant, yet it yields the optimal outcomes on the METEOR metric.
 Critically, all our model variants maintain a human fluency score above 4.3 (out of 5), demonstrating that the fluency requirement is not compromised. This is not surprising, given the impressive ability of current language models to generate fluent texts.

\subsection{Distilled Models}
From \autoref{tab:rouge-results} we observe a noteworthy finding: when trained on the GPT-4-generated data, Flan-T5\textsubscript{H} surpasses all the non-distilled alternatives, which were trained on the original CTR data. Notably, it also surpasses the GPT-4 model, where the latter is prompted in the few-shot setting, just as it was when generating the training data for Flan-T5\textsubscript{H}.
Given the inherent limitation of GPT-4 in that it is not open-sourced and cannot be fine-tuned on our data, this puts it at a distinct disadvantage compared to the other models, thereby obstructing its broader application.

We also note that Flan-T5\textsubscript{H} appears to further benefit from both the auxiliary highlights-focused RL finetuning and the incorporation of the highlights-attentive decoding strategy.
We find that the decoding strategy outperforms the RL approach based on ROUGE and BertScore metrics, similar to the findings for the non-distilled models, while the RL approach appears to be more effective on the METEOR metric. We also note that the combination of both strategies does not result in any additional advantages. 
Ultimately, just as in the non-distilled setting, we find that the coherency of the generated outputs remains uncompromised.
In total, the combination of the highlights-focused strategies with the improved GPT-4-generated data leads to significant improvements, outperforming the baseline model by over 30 ROUGE-L points, and resulting in a reliable and effective CTR model for future downstream use.


\subsection{Distillation Data Quality Assessment}\label{subsec:data_quality}
To shed light on the distillation success, we analyze the quality of the generated dataset.
To that end, we first sample 10 GPT-4-generated instances and manually identify their corresponding highlights, which we treat as ``gold'' highlights.
Then, we apply \citet{slobodkin-etal-2022-controlled}'s automated silver annotation methodology
 (see \S\ref{sec:Background}) on the GPT-4-generated summaries, leading to a new set of highlights for each pair of input text and GPT-4-generated summary. These highlights, combined with their corresponding inputs and summaries, represent the original CTR dataset. Alternatively, the original highlights presented to GPT-4, combined with the input texts and the summaries it generated, represent the new dataset.
 Subsequently, we proceed to calculate the ROUGE-L F1 score between the manually annotated highlights and each of the automatically-generated variants.
Our analyses reveal that the GPT-4-generated instances demonstrate a significantly greater alignment with the manually-annotated instances than those from the original dataset, achieving a ROUGE-L score of $79\%$, as opposed to $67.9\%$.


\begin{table}[]
\centering
\resizebox{\columnwidth}{!}{%
\begin{tabular}{lccl}
\hline
                                                               & Faithfulness (P) & Coverage (R) & F-1  \\ \hline
LED\textsubscript{H}                                           & 65.8             & 72.9         & 69.2 \\
Flan-T5\textsubscript{H}                                       & 71.1             & 74.0         & 72.5 \\
Flan-T5\textsubscript{H} (distil.)                             & 79.1             & 90.8         & 84.6 \\
\hspace{3mm} {\small+ RL}                                                            & 81.3             & \textbf{93.4}         & 86.9 \\
\hspace{3mm} {\small+ Con. Decoding}                                                & \textbf{85.6}             & 91.3         & \textbf{88.3} \\
\begin{tabular}[c]{@{}l@{}} \hspace{3mm} \vspace{-1.5mm}  {\small+ RL}\\  \hspace{6mm}  {\small+ Con. Decoding}\end{tabular} & 83.4             & 92.3         & 87.7 \\ \hline
\end{tabular}%
}
\caption{Fact-wise faithfulness (P), coverage (R), and F-1 scores between generated summaries and the highlighted spans for LED\textsubscript{H}, Flan-T5\textsubscript{H}, and four variants of the distilled Flan-T5\textsubscript{H}: regular, with the RL finetuning ("+RL"), with the controlled decoding approach ("+ Con. Decoding") and with their combination. For each metric, the best models are in bold.}
\label{tab:pyramid_comparison}
\end{table}

\subsection{Manual Performance Analysis}\label{subsec:performance_analysis}
To further evaluate the effectiveness of various components within our approach, we adopt the manual analysis methodology proposed in the CTR paper~\cite{slobodkin-etal-2022-controlled}. Consistent with the original authors' procedure,  we select 10 random samples from the test set.
Subsequently, we compute precision, recall, and F-1 scores for highlighted information units for a selection of six models utilized in this study: LED\textsubscript{H}, Flan-T5\textsubscript{H}, the distilled Flan-T5\textsubscript{H}, and the three variants of the distilled Flan-T5\textsubscript{H} equipped only with RL training, only with the controlled decoding, and with both approaches, respectively.\footnote{For further details, refer to Appendix~\ref{sec:performance_analysis_settings}.}
Our calculations cover 195 highlighted input units and approximately 180 system summary units for each model. 
The results of this analysis are illustrated in Table \ref{tab:pyramid_comparison}.

We observe that Flan-T5\textsubscript{H} exhibits significantly increased faithfulness to the highlights compared to LED\textsubscript{H}, with a modest improvement in highlight coverage.
Conversely, training on the GPT-4-generated data markedly augments adherence to the highlights as well as leading to significant highlight coverage, thus demonstrating the approach's inherent value.

Our findings also suggest that the application of RL finetuning primarily improves highlight coverage. The seemingly modest contribution to faithfulness could potentially be attributed to the RL mechanism's attempt to achieve a balance between reward optimization and coherence-preservation (due to the KL-divergence term), a requirement that might necessitate incorporating additional information from the surrounding context.
Alternatively, the deployment of highlight-centric controlled decoding displays a more pronounced impact on adherence to highlights. This phenomenon could potentially be attributed to the strategy's more rigorous enforcement of highlight constraints, thereby enforcing limited deviation from the highlights. The modest advancement in coverage could be attributed to the lookahead mechanism of controlled decoding.
At every generation step, the algorithm favors tokens whose 'greedy' completion leads to better highlight adherence, rather than exploring multiple potential trajectories for each candidate akin to a k-beam style. In doing so, this method inadvertently neglects additional, more suited candidates, whom a beam search could capture and thereby enhance coverage.
Lastly, the combination of both strategies results in a mid-point performance between the individual gains attained by each strategy in both faithfulness and coverage. These findings indicate that while merging the strategies does enhance the weaker model's performance in each metric, it also has a trade-off effect on the other metric. Future research will explore methods to harness the complementary aspects of these strategies while mitigating this trade-off effect.

\section{Discussion and Future Work}\label{sec:Discussion}
Our empirical assessments reveal the merits of each of the proposed methods. Specifically, the utility of the distillation process emerges as complementary to both reinforcement learning (RL) training and controlled decoding. 
The latter methods still have room to play in both enforcing the content preservation constraints beyond the sheer training data, as well as in overcoming a certain level of noise that persists also in the GPT-4 generated data (as shown in \S\ref{subsec:data_quality}).

Conversely, the combination of controlled decoding and highlight-centric RL approach did not yield improvements over the better-performing method (controlled decoding, in our experiments). Given these strategies impact different stages of the generation process, namely training and inference, they may possess the potential for synergy. For example, the decoding strategy may be integrated into the RL training's sampling phase, possibly increasing the model's awareness of the decoding constraints during training.
Our performance analysis, detailed in \S\ref{subsec:performance_analysis}, further supports this hypothesis by demonstrating that each technique amplifies different aspects of the highlights: one notably improves their coverage while the other augments adherence, suggesting the exploration of this potential synergy in future research.

Furthermore, the existing implementation of highlight-centric controlled decoding is computationally demanding, given its requirement to generate a complete completion for every candidate token at each generation step. \citet{wan-etal-2023-faithfulness} motivated the need to generate entire summaries by the expectation of current faithfulness metrics for full summary inputs. Yet, as our method does not employ faithfulness metrics, it would be insightful to explore how to leverage partial summaries, rather than complete ones, and thus to significantly reduce the computational overhead. 

 Finally, given the high-quality highlight adherence and coverage exhibited by our best models, investigating their incorporation into modular summarization pipelines emerges as a promising research direction. 
 We suggest exploring this direction in future research.

\section{Conclusion}\label{sec:Conclusion}
In this study, we addressed the lack of a high-quality Controlled Text Reduction (CTR) model by focusing on two pivotal aspects: the amplification of the highlight signals and the mitigation of noise within the training data.
We started by proposing two distinct strategies aiming at augmenting the highlights signal. 
The first strategy emphasized this signal during training, where we combined an RL approach with a custom highlights-oriented reward.  The second strategy was introduced during inference, where we employed a controlled decoding mechanism that prioritizes generation paths ensuring higher adherence to the highlights.
Furthermore, we addressed the intrinsic noise in the CTR dataset by generating new instances using GPT-4, significantly enhancing the dataset quality.

Empirical evidence shows the effectiveness of our proposed methodology. Each of our highlight-centric strategies individually led to significant improvements over the baseline model in terms of highlight-matching capabilities.
Additionally, training on the GPT-4-generated data yielded further improvements, outperforming each of the non-distilled variants, trained on the original CTR dataset.  In total, our highest-performing models achieved state-of-the-art results, outperforming the baseline by more than $30$ ROUGE-L points, while also preserving comparable levels of coherency.
Future work would focus on improving the combination and efficiency of the different components, as well as on the incorporation of our best-performing models in modular summarization pipelines.


\section{Limitations}
Although the risk involved in our work is minimal, like other advanced language generation models available today, we cannot ensure that our model will consistently produce accurate information, despite our attempts to utilize clean highlighted data and controlled decoding. Hence, it is crucial to exercise caution when employing the model in real-world scenarios and thoroughly test it before deploying it.

In addition, the controlled decoding method we use is limited by its slow speed, which may hinder its practicality for production systems. Additionally, the rigid constraints imposed during decoding can restrict the model's linguistic flexibility and creative potential, potentially limiting its suitability for generating varied or innovative outputs. These considerations highlight the need to carefully evaluate the trade-offs before implementing controlled decoding in production environments, and hence should be further studied in future research.

\section*{Acknowledgements}
This work was supported by the Israel Science Foundation (grant no. 2827/21), and a grant from the Israel Ministry of Science and Technology.

\bibliography{anthology,custom}
\bibliographystyle{acl_natbib}
\appendix

\section{Reward Tuning}\label{sec:reward_tuning}
We train the modified dual-reward  Quark model with three pairs of highlights-oriented ROUGE rewards: ROUGE Precision+ROUGE F1, ROUGE Recall+ROUGE F1, and ROUGE Precision + ROUGE Recall. From \autoref{tab:rouge_rewards_tuning_Quark}, which shows the ROUGE scores on the CTR development set, it arises that alternating between ROUGE precision and ROUGE recall rewards yields the best results. 


\begin{table}[ht!]
\centering
\resizebox{\columnwidth}{!}{%
\begin{tabular}{lccccc}
\hline
R-L Rewards & R-1 & R-2 & R-L & M & \multicolumn{1}{l}{BertScore} \\ \hline
Precision + F1 & 74.0 & 60.7 & 67.3 & 69.8 & 76.3 \\
Recall + F1 & 74.1 & 60.9 & 67.6 & 70.4 & 76.2 \\
Only F1 & 74.4 & 61.2 & 67.3 & 70.0 & 76.4 \\
Precision + Recall & \textbf{76.7} & \textbf{64.0} & \textbf{70.4} & \textbf{74.0} & \textbf{77.7} \\ \hline
\end{tabular}%
}
\caption{ROUGE, METEOR (M), and BertScore results of the Flan-T5\textsubscript{H} model, combined with the RL strategy, on the CTR development set. We experimented with different pairs of alternating highlights-oriented ROUGE-L rewards, as well as only a single ROUGE-L F1 reward.}
\label{tab:rouge_rewards_tuning_Quark}
\end{table}

\begin{figure}[ht!]
    \centering
    \includegraphics[width=0.55\columnwidth]{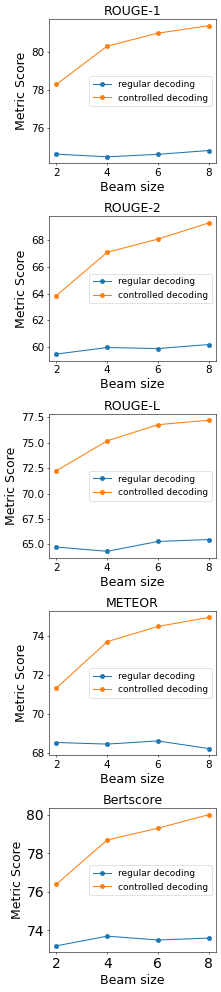}
    \caption{The scores of fine-tuned Flan-T5\textsubscript{H} with controlled decoding compared to regular decoding at various beam sizes. We present a comparison of the generated summary to the highlights concatenation.}
\label{fig:controlled_decoding_beam_tuning}
\end{figure}


\begin{table}[ht!]
\centering
\resizebox{\columnwidth}{!}{%
\begin{tabular}{lccccc}
\hline
 & \multicolumn{1}{l}{R-1} & \multicolumn{1}{l}{R-2} & \multicolumn{1}{l}{R-L} & M & \multicolumn{1}{l}{BertScore} \\ \hline
METEOR & 77.0 & 60.3 & 61.7 & \textbf{71.3} & 74.3 \\
ROUGE-L & \textbf{78.3} & \textbf{63.9} & \textbf{72.3} & \textbf{71.3} & \textbf{76.4} \\ \hline
\end{tabular}%
}
\caption{F1 results for ROUGE, METEOR (M), and BertScore compared to the concatenation of the highlights on the dev set. The results are for Flan-T5\textsubscript{H} with a beam size of 2, incorporating two variants of the highlights-centric score during decoding: METEOR (F1) compared to the highlights, and ROUGE-L (F1) compared to the highlights. \textbf{Bold} marks the highest results on each metric.}
\label{tab:controlled_decoding_score_tuning}
\end{table}

\section{Highlights-Focus Controlled Decoding Hyperparameter Tuning}
\label{sec:Highlights-Focus Controlled Decoding Hyperparameter Tuning}
For our highlights-oriented decoding strategy, we experiment with two scores: METEOR score and ROUGE-L F1 score. These are juxtaposed against the concatenated highlights to assess the potential completion's adherence to the highlights. \autoref{tab:controlled_decoding_score_tuning} shows the results on the CTR development set. We note that the ROUGE-L score consistently offers superior or at least equivalent performance in comparison to the METEOR score. Therefore, based on these findings, we make the informed decision to adopt the ROUGE-L score as our primary evaluation metric in all subsequent experiments.
Furthermore, we undertake a series of experiments with varied beam sizes encompassing 2, 4, 6, and 8. The outcomes, as observed on the development set, are illustrated in \autoref{fig:controlled_decoding_beam_tuning}. Corroborating the findings of \citet{wan-etal-2023-faithfulness}, we discern a marginal impact of the increase in beam size on regular decoding. However, its influence on the highlight-focused controlled decoding escalates in parallel with the beam size across all metrics. In light of these findings, we opt for a beam size of 8 for the remainder of our experiments.

\section{GPT-4 Prompt}\label{sec:GPT-4_Prompt}
For deploying GPT-4, we use a prompt that consists of some basic instructions, as well as two exemplars.
The in-context examples consist of a modular generation, where we separate the task into three steps: (1) highlights extraction and enumeration, (2) highlights consolidation sentence-by-sentence, and (3) generation of the final reduction (see \autoref{fig:GPT4_distillation_prompt}).
Additionally, we already perform the highlights extraction step for the current example, leaving only the consolidation and generation of the final reduction to the model.

\section{Prompt-Tuning}\label{sec:prompt_tuning}
To improve the quality of the CTR trainset, we test both GPT-4 and GPT-3.5 (ChatGPT), both with modular prompting and with regular prompting, where each exemplar's answer consists solely of the final reduction.
From \autoref{tab:chatgpt_gpt4_prompt_tuning}, which shows the ROUGE and METEOR scores on 50 instances of the CTR development set, we observe that indeed the modular approach substantially improves both models' performances and that GPT-4 is superior to ChatGPT in handling the task.
Additionally, from a manual analysis of GPT-4's generations, we find that occasionally it misses a highlighted span in the highlights-listing step, leading to its absence from the final summary. 
Consequently, we also experiment with a prompt that already incorporates the highlights-listing step, leaving only their consolidation and the generation of the final reduction to the model. This version indeed which improves the model's performance (see
\autoref{tab:gpt4_prompt_highlights_incorporation_tuning}), making it the final version we use.


\begin{table}[]
\centering
\resizebox{\columnwidth}{!}{%
\begin{tabular}{llccccc}
\hline
Model &  & R-1 & R-2 & R-L & M & \multicolumn{1}{l}{BertScore} \\ \hline
GPT-3.5 & regular & { 59.9} & { 40.7} & { 47.7} & { 49.7} & 70.0 \\
 & modular & { 72.7} & { 61.1} & { 68.1} & { 75.0} & 75.7 \\ \hline
GPT-4 & regular & { 71.0} & { 55.1} & { 64.5} & { 63.1} & 76.4 \\
 & modular & { \textbf{81.0}} & { \textbf{66.0}} & { \textbf{73.0}} & { \textbf{76.3}} & \textbf{80.5} \\ \hline
\end{tabular}%
}
\caption{ROUGE, METEOR (M), and BertScore results on 50 instances from the CTR development set of GPT-3.5 and GPT-4 models. For each model, we prepend an exemplar to the prompt, in two fashions: "regular", where the exemplar simply generated the custom summary; and "modular", where the exemplar first performs the intermediate sentence-by-sentence highlights consolidation, before generating the final summary.}
\label{tab:chatgpt_gpt4_prompt_tuning}
\end{table}

\begin{table}[]
\centering
\resizebox{\columnwidth}{!}{%
\begin{tabular}{cccccc}
\hline
 & R-1 & R-2 & R-L & M & \multicolumn{1}{l}{BertScore} \\ \hline
w\textbackslash{}o list\textsubscript{h} & { 84.3} & { 68.4} & { 76.4} & { 81.4} & \textbf{81.8} \\
\multicolumn{1}{l}{with list\textsubscript{h}} & \multicolumn{1}{r}{{ \textbf{84.8}}} & \multicolumn{1}{r}{{ \textbf{71.3}}} & \multicolumn{1}{r}{{ \textbf{77.1}}} & { \textbf{83.6}} & \textbf{81.8} \\ \hline
\end{tabular}%
}
\caption{ROUGE, METEOR and BertScore results, compared to the concatenated highlights, on 50 instances from the CTR development set of GPT-4 models, once when the prompt does not contain highlights enumeration of the current instance ("w\textbackslash{}o list\textsubscript{h}") and once when it does contain it ("with list\textsubscript{h}").}
\label{tab:gpt4_prompt_highlights_incorporation_tuning}
\end{table}



\section{Qualitative Analysis}\label{sec:qualitative_analysis}
\begin{figure*}[t!]
\centering
    \includegraphics[width=1\linewidth]{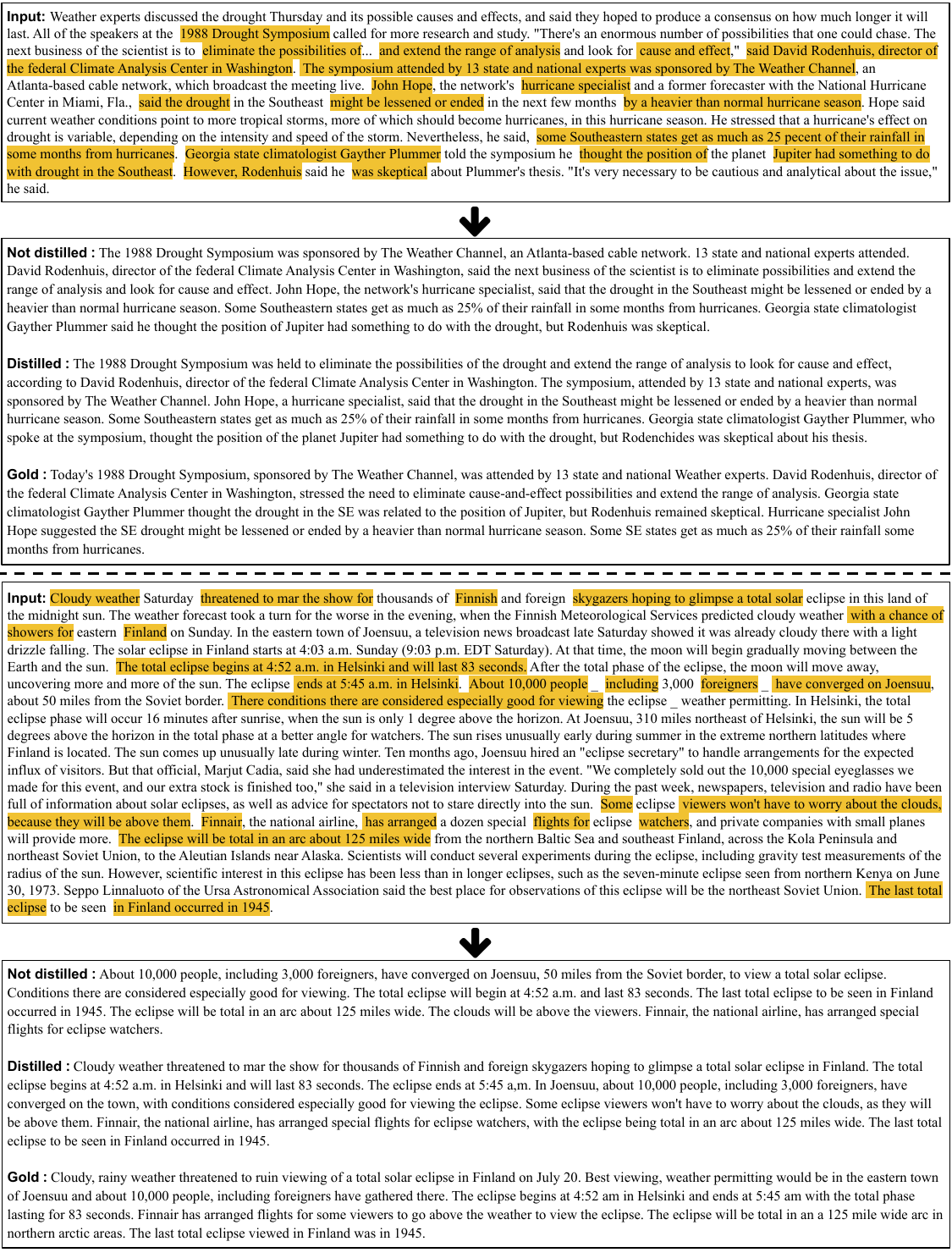}
    \caption{Examples including the input, gold output and predictions with various baselines.}
    \label{fig:qualitative_analysis}
\end{figure*}

\autoref{fig:qualitative_analysis} exhibits two instances from the test set, consisting of input texts along with their corresponding highlights. Accompanying each instance are the outputs generated by the non-distilled Flan-T5\textsubscript{H}, the output generated by the distilled Flan-T5\textsubscript{H}, and the original gold summaries.

In the first example, the non-distilled model achieved a ROUGE-L score of 58.6 when compared to the gold summary, while its distilled counterpart received a score of 46.5. Despite the apparent superiority of the non-distilled model, we note that while it did cover all the highlighted spans, it also added several non-highlighted segments (``an Atlanta-based cable network'', ``the next business of the scientist'', ``the network's'' and ``look for'' in the context of looking for cause and effect). In contrast, the distilled model, besides capturing all highlighted content, only added ``look for''. Consequently, the distilled model delivered a more fitting output, contradicting its supposed deficiency in terms of the ROUGE-L metric relative to the gold summary. 
Alternatively, when compared to the concatenated highlights, which yielded scores of 68.8 for the non-distilled model, and 78.9 for the distilled model, the results are more reflective of the actual performance.

In the second example, the distilled model amounts to 83.3 in comparison to the concatenated highlights, as opposed to a mere 43.2 when compared to the gold summary. Interestingly, despite the fact that its output nearly covers all the highlighted spans (except for ``with a chance of showers for...Finland'' and ``in Helsinki'' in the context of where the eclipse would end), and introduces very minimal non-highlighted information (``thousands of'', ``3,000'' and ``the national airline''), it still results in a lower ROUGE-L score when aligned with the gold summary. 
This discrepancy can be attributed to two significant factors. Firstly, the distinction in abstraction techniques employed between the gold summary and the generated output can contribute to the reduced scores. Secondly, the gold summary introduces information that is absent from the original input text, and therefore from the highlights (``on July 20''). Both these elements can explain the divergence between the model's performance as per the ROUGE-L metric and its actual competency.
This discrepancy can be resolved when directly comparing the generated summary to the highlights, which offers a more accurate reflection of its quality.

\section{Fluency Human Annotation Protocol}\label{sec:Fluency_Human_Annotation_Protocol}
\begin{figure*}[t!]
    \includegraphics[width=1\linewidth]{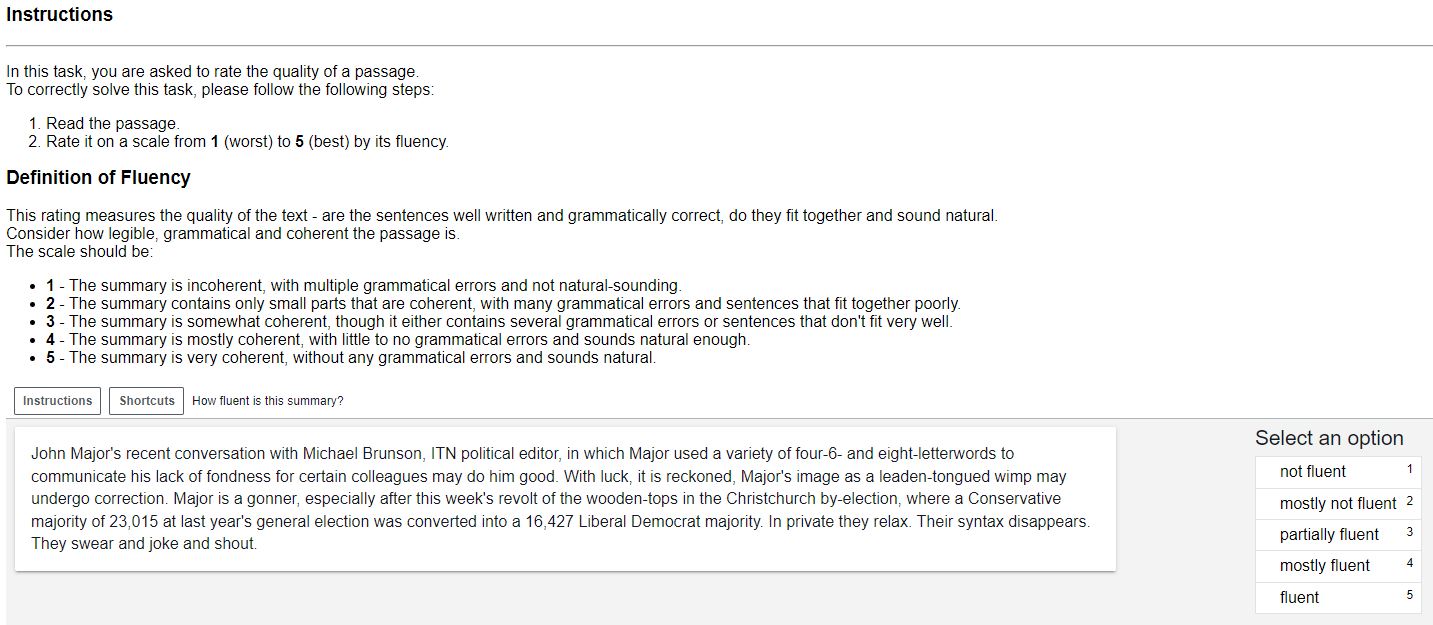}
    \caption{Example of the data collection interface used by the crowd-workers to evaluate the fluency of summaries.}
    \label{fig:fluency_evaluation_interface}
\end{figure*}

We ask crowd-workers to rate the fluency of the texts generated by the baseline supervised model, the PPO model and the dual-reward Quark model. 
Our group of crowd-workers consists of reliable workers that have shown a good understanding of different semantic tasks including summarization in previous experiments. To evaluate, we randomly select 100 documents from our test set and evaluate their corresponding generated text by the three aforementioned models (300 samples in total). We design a simple Amazon Mechanical Turk interface, where we present each time one of the 300 samples (see \autoref{fig:fluency_evaluation_interface}). Following \citet{slobodkin-etal-2022-controlled} we use a 5-point Likert scale to evaluate the fluency of the generated summaries. Additionally, we add criteria explaining each score, to reduce ambiguity and ensure consistent ratings (see \autoref{fig:fluency_evaluation_interface}). Assessing an average response period of 30 seconds, we priced each response with 10 \textcentoldstyle.

\begin{figure*}[t]

\lstdefinestyle{promptStyle}
{
    basicstyle={\footnotesize\ttfamily},
    numbers=left,numberstyle=\footnotesize,
    xleftmargin=2.8em,
    xrightmargin=1.5em,
    showstringspaces=false,
      showspaces=false,
        showtabs=false,
    tabsize=2,
    breaklines=true,
        flexiblecolumns=true,
        escapeinside={<@}{@>},
          breakatwhitespace=true
}

\newtcblisting{mylisting}[1]{
  enhanced,
  listing only,
  boxrule=0.8pt,
  sharp corners=downhill,
  top=0mm,
  bottom=0mm,
  left=2mm,
  right=0mm,
  boxsep=0mm,
  colframe=black,
  colback=white,
  listing options={
    style=#1
  }
}

\definecolor{instructionsColor}{rgb}{0.1, 0.5, 0.1}

\begin{mylisting}{promptStyle}
<@\textcolor{instructionsColor}{In this task, you are presented with a passage, where some parts are "highlighted" (namely, there are <highlight\_start> and <highlight\_end> tokens before and after each such span). Your job is to generate a summary that covers all and only the "highlighted" spans.}@>

<@\color{red}Example1:@>
<@\color{blue}Passage:@> <highlight_start>Ben Johnson spent his homecoming<highlight_end> in seclusion, <highlight_start>without the<highlight_end> Olympic <highlight_start>gold medal and<highlight_end> the <highlight_start>hero's welcome<highlight_end>, as Canadians bemoaned the fate of the sprinter who failed the drug test.
...

<@\color{blue}Answer: The highlighted spans are:@>
 1. Ben Johnson spent his homecoming
...
<@\color{blue}The highlights spans are combined as follows:@>
Spans 1,2,3,4 are combined to form sentence 1: Ben Johnson spent his homecoming without the gold medal and hero's welcome. 
...
<@\color{blue}So, the answer is:@>
Ben Johnson spent his homecoming without the gold medal and hero's welcome.
...

<@\color{red}Example2:@>
<@\color{blue}Passage:@> Japanese writer <highlight_start>Kazuo Ishiguro won the<highlight_end> 1989 <highlight_start>Booker Prize<highlight_end>, Britain's top literary award, <highlight_start>for his novel "The Remains of the Day<highlight_end>," judges <highlight_start>announced Thursday<highlight_end>.
...

<@\color{blue}Answer: The highlighted spans are:@>
 1. Kazuo Ishiguro won the
...
<@\color{blue}The highlights spans are combined as follows:@>
Spans 1,2,3,4 are combined to form sentence 1: It was announced Thursday that Kazuo Ishiguro won the Booker Prize for his novel "The Remains of The Day". 
...
<@\color{blue}So, the answer is:@>
It was announced Thursday that Kazuo Ishiguro won the Booker Prize for his novel "The Remains of The Day".
...

<@\textcolor{red}{Now your turn:}@>
<@\color{blue}Passage:@>  <highlight_start>The motion picture industry's most coveted award<highlight_end>, <highlight_start>Oscar<highlight_end>, <highlight_start>was created 60 years ago and 1,816 of the statuettes have been produced so far.<highlight_end>
...
<@\color{blue}Answer: The highlighted spans are:@>
 1. The motion picture industry's most coveted award
...
<@\color{blue}The highlights spans are combined as follows:@>
\end{mylisting}
\caption{Example prompt provided to GPT-4. The prompt consists of basic instructions, two in-context examples, and the instance input. The examples demonstrate a modular pipeline, where we first extract the highlights, then consolidate them sentence-by-sentence, and lastly generate the final reduction.}
\label{fig:GPT4_distillation_prompt}
\end{figure*}

\begin{table*}[]
\centering
\begin{tabular}{|ll|ccccc|}
\hline
\multicolumn{2}{|l|}{model}                                                                                                                                                                       & R-1               & R-2               & R-L               & M                 & Bertscore         \\ \hline
\multicolumn{1}{|l|}{\multirow{4}{*}{\rotatebox{90}{{\small \hspace{-5mm} Non-Distilled}}}} & LED\textsubscript{H}                                  & 70.2              & 53.8              & 54.4              & 64.3              & 71.9              \\
\multicolumn{1}{|l|}{}                                                                                                   & \hspace{3mm}  {\small+ RL}                                                                   & 70.8              & 56.1              & 56.5              & \textbf{70.9}     & 73.7              \\
\multicolumn{1}{|l|}{}                                                                                                   & \hspace{3mm}  {\small+ Con. Decoding}                                                        & \textbf{77.4}     & \textbf{64.7}     & \textbf{71.9}     & 69.5              & \textbf{77.8}     \\
\multicolumn{1}{|l|}{}                                                                                                   & \begin{tabular}[c]{@{}l@{}} \hspace{3mm} \vspace{-1.5mm}  {\small+ RL}\\  \hspace{6mm}  {\small+ Con. Decoding} \end{tabular}  & 75.4              & 60.9              & 67.5              & 68.6              & 76.7              \\ \hline
\multicolumn{1}{|l|}{\multirow{4}{*}{\rotatebox{90}{{\small \hspace{-5mm} Distilled}}}}                 & LED\textsubscript{H}                                  & 77.9              & 64.0              & 69.7              & 75.6              & 74.0              \\
\multicolumn{1}{|l|}{}                                                                                                   & \hspace{3mm}  {\small+ RL}                                                                   & 76.6              & 61.7              & 66.8              & 75.4              & \hspace{0.8mm} \textbf{78.2$^*$} \\
\multicolumn{1}{|l|}{}                                                                                                   & \hspace{3mm}  {\small+ Con. Decoding}                                                        & \hspace{0.8mm} \textbf{81.4$^*$} & \hspace{0.7mm} \textbf{68.5$^*$} & \hspace{0.8mm} \textbf{77.9$^*$} & \hspace{0.8mm} \textbf{75.5$^*$} & 77.1              \\
\multicolumn{1}{|l|}{}                                                                                                   & \begin{tabular}[c]{@{}l@{}} \hspace{3mm} \vspace{-1.5mm}  {\small+ RL}\\  \hspace{6mm}  {\small+ Con. Decoding}\end{tabular} & 78.1              & 63.8              & 73.0              & 73.1              & 75.6              \\ \hline
\end{tabular}
\caption{ROUGE, METEOR, and Bertscore results on the CTR testset, compared to the concatenated highlights, of all the different variants tested in this work, with LED\textsubscript{H} as the backbone model. In addition to the baseline LED\textsubscript{H}, we also evaluate the combination of LED\textsubscript{H} with the highlights-sensitive decoding strategy ("+ Con. Decoding"), with the highlights-focused RL strategy ("+ RL"), and with their combination. We also evaluate all those variants when trained with the GPT-4-distilled trainset ("Distilled"). For each metric, The best non-distilled and distilled LED\textsubscript{H} models are in bold, with the best overall model having an asterisk.}
\label{tab:rouge-results-LED}
\end{table*}

\section{Results of All Variants with LED\textsubscript{H} as the Backbone Model} \label{sec:Results of All Variants with LED as the Backbone}
\autoref{tab:rouge-results-LED} shows performance results for all the variants explored in this work, with LED\textsubscript{H} as the backbone model.

\section{Performance Analysis Settings}\label{sec:performance_analysis_settings}
To further examine the efficiency of our different components, we follow \citet{slobodkin-etal-2022-controlled} and
manually assess several of our models on two levels: (1) precision to the highlighted content and (2) recall of the highlighted spans. 
For that, we compare each system summary span to the source highlights.
To that end, we randomly select 10 samples from our test set, with their corresponding system summaries (one for each of the models - 50 in total).
Then, following the notion of Summary Content Unit (SCU) in the Pyramid method for summarization evaluation \citep{nenkova-passonneau-2004-evaluating}, we extract such units from both the summary and the source highlighted spans using the Summary Evaluation Environment (SEE) interface, described in their paper. 

Then, to calculate the precision, for each summary unit, we manually search for a matched highlighted unit conveying the same information, to determine whether the summary unit is mentioned in the highlights (TP) or not (FP), and then calculate the (micro-)precision to the highlighted content.

For the recall calculations, we also count the number of False Negative (FN) summary facts, compared to the facts in the highlights, namely highlighted information units that were absent from the system summaries. Then, combined with the TP count, we calculate the (micro-)recall of the highlighted content.

\end{document}